\documentclass[sn-aps]{sn-jnl}

\usepackage{graphicx}%
\usepackage{multirow}%
\usepackage{amsmath,amssymb,amsfonts}%
\usepackage{amsthm}%
\usepackage{mathrsfs}%
\usepackage[title]{appendix}%
\usepackage{xcolor}%
\usepackage{textcomp}%
\usepackage{manyfoot}%
\usepackage{booktabs}%
\usepackage{algorithm}%
\usepackage{algorithmicx}%
\usepackage{algpseudocode}%
\usepackage{listings}%
\usepackage{subcaption}
\usepackage{numprint}

\theoremstyle{thmstyleone}%
\theoremstyle{thmstyletwo}%

\theoremstyle{thmstylethree}%
\newcommand{\blockcomment}[1]{}
\begin{document}

\title[Article Title]{Unsupervised Welding Defect Detection Using Audio and Video}


\author[1]{\fnm{Georg} \sur{Stemmer}}
\author*[1]{\fnm{Jose A.} \sur{Lopez}}\email{jose.a.lopez@intel.com}
\author[1]{\fnm{Juan A.} \sur{Del Hoyo Ontiveros}}
\author[1]{\fnm{Arvind} \sur{Raju}}
\author[1]{\fnm{Tara} \sur{Thimmanaik}}
\author[1]{\fnm{Sovan} \sur{Biswas}}


\affil*[1]{\orgname{Intel Corp.}, \orgaddress{\street{2200 Mission College Blvd.}, \city{Santa Clara}, \postcode{95054}, \state{CA}, \country{USA}}}


\abstract{In this work we explore the application of AI to robotic welding. Robotic welding is a widely used technology in many industries, but  robots currently do not have the capability to detect welding defects which get introduced due to various reasons in the welding process. We describe how deep-learning methods can be applied to detect weld defects in real-time by recording the welding process with microphones and a camera. Our findings are based on a large database with more than 4000 welding samples we collected which covers different weld types, materials and various defect categories. All deep learning models are trained in an unsupervised fashion because the space of possible defects is large and the defects in our data may contain biases. We demonstrate that a reliable real-time detection of most categories of weld defects is feasible both from audio and video, with improvements achieved by combining both modalities. Specifically, the multi-modal approach achieves an average Area-under-ROC-Curve (AUC) of 0.92 over all eleven defect types in our data. We conclude the paper with an analysis of the results by defect type and a discussion of future work.}




\maketitle

\section{Introduction}\label{sec1}
Robotic arc welding, i.e., the use of robots for automating the arc welding process, is a key manufacturing technology in many industries. As the quality of a weld depends on many factors, even a robot that repeats each step of the process perfectly will produce defective welds from time to time. The time at which a defect gets detected in the welding process has direct impact on the cost associated with correcting the problem. So to minimize the overall defects and reduce the correction cost there is a growing interest in detecting anomalies in real-time, i.e., during the welding process, rather than post weld defect detection. Ideally, a future intelligent manufacturing system would be able to adjust the welding robot’s parameters automatically even before the failure starts to occur in the welding process.

In this work, we investigate the use of deep learning models for automatic weld defect detection in real-time using a camera and microphones. Cameras can monitor the weld pool geometry and oscillations, which are known to be predictive of weld defects \cite{yu_2023, ma_2021, zou_2020}. Cameras need to have line-of-sight to the weld pool, and this consequently limits the mounting options. Microphones, on the other hand, can capture audible disturbances in the welding process without line-of-sight. They have been shown to provide useful information about defects \cite{ji_2023,alcaraz_2023,chen_2023,Madhvacharyula_2022,na_2021,lv_2017,sumesh_2015}.
 We investigate in the experimental section for several defect types whether they can be better detected visually or acoustically, and how the two modalities perform in combination.
 
Current deep learning models require large amounts of training data to estimate their parameters. We address this issue by collecting more than 4000 samples of good and defective welds, which is, to best of our knowledge, significantly more than what has been reported for similar datasets in the literature. Still, we consider the number of samples too small to train a classifier that can distinguish each relevant defect type reliably, given the variation of the input signals that can be expected in a real application. Therefore we decided to address the weld defect detection problem with an unsupervised anomaly detection approach. Anomaly detection models are trained on good welds only: all defects in our dataset occur solely during evaluation and are unseen in training. While this makes the defect specific performance more difficult to tune, we believe that the resulting performance is more representative of a real application.

Of course, we are not the first interested in bringing the benefits of AI to established spaces like manufacturing, and welding in particular. Companies have produced AI-based anomaly detection solutions in recent years \cite{realitycheck_ad, realitycheck_ad_presentation, amazon_monitron, xiris_ai_product}, and  numerous authors have explored the use of modern data-driven algorithms to improve weld defect detection \cite{yu_2023, eren_2023, breitenbach_2021}, which can be costly to remediate \cite{meyer_2022}. In \cite{asif_2022}, the authors used sequence tagging and logistic regression to detect welding defects. Mohanasundari et al. \cite{mohanasundari_2021}  used post-weld images to classify defects. Buongiorno et al. \cite{buongiorno_2022} leveraged thermographic image sequences from an infrared camera to detect defects.  
Our work is also related to predictive maintenance solutions, which tend to use vibration or acoustic-emission sensors. Acoustic emission sensors have even been used successfully to monitor weld quality \cite{asif_2022} as well. There is good reason for using this type of sensors for machine condition monitoring, as 90\% of rotating machinery uses rolling-element bearings, which tend to be points of failure \cite{graney_and_starry_2012}. On the other hand, welding is a much more complex process. Therefore we expect that microphones and cameras are better suited for detecting weld defects than acoustic emission or vibration sensors.

The main contributions of this work are the following: We collect a large multi-modal dataset of samples of robotic arc welding in a real industrial environment. It covers different weld types, various welding parameter configurations, and steel types. The size of the dataset allows us to train deep ML models, as opposed to shallow or analytical models often described in the literature (e.g., \cite{asif_2022}). With these deep ML models we are able to demonstrate that camera and microphone are adequate sensors for real-time weld defect detection. Our unsupervised approach ensures that the model is not biased to defect characteristics which are specific for our data collection setup and allows us to compare results between different defect types and and modalities. Finally, we demonstrate that defect detection performance can be improved using a multi-modal combination of both sensor types.

In the next sections we provide a comprehensive description of how the dataset has been collected. We introduce our experimental setup, quality metrics and ML models. The performance of unsupervised algorithms trained in single and multi-modal fashion on this dataset is evaluated experimentally and we present our conclusions.
 
\nocite{*}

\section{Data Set}
\label{sec:data}
The goal of our data collection was to record enough samples for each of the most important weld categories to make statistically valid comparisons across modalities and weld defect types. We collaborated with a supplier that has access to automotive factories to conduct the data acquisition in a real factory environment. The welds were generated using a 6-axis arc welding robot using two steel types and thicknesses. The steel types were selected to be often-used varieties for automotive applications in India. The first type is known as ``FE410'' and the second ``BSK46'' type has higher carbon content and is used for higher-strength applications. The thicknesses used were 7mm for most of the samples and 3mm for specific defects that could not be efficiently induced using the 7mm steel, like burnthrough.

\subsection{Recording Setup, Data Collection Procedure, and Limitations}

\begin{figure}[ht]
     \centering
     \begin{subfigure}[b]{\textwidth}
         \centering
         \includegraphics[width=\textwidth]{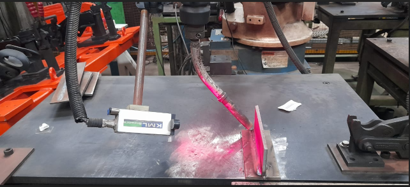}
         \caption{KML camera.}
         \label{fig:kml_camera}
     \end{subfigure}
     \vfill
     \begin{subfigure}[b]{0.4\textwidth}
         \centering
         \includegraphics[width=\textwidth]{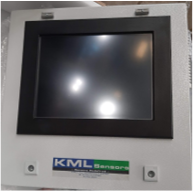}
         \caption{KML control panel.}
         \label{fig:kml_control_panel}
     \end{subfigure}
     \hfill
     \begin{subfigure}[b]{0.4\textwidth}
         \centering
         \includegraphics[width=\textwidth]{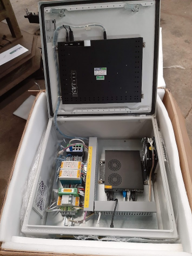}
         \caption{KML computer.}
         \label{fig:kml_computer}
     \end{subfigure}	
        \caption{KML camera and computer.}
        \label{fig:kml_camera_and_computer}
\end{figure}

The data collection station comprised an AII-V6 6-axis arc welding robot \cite{otc_v6}, an Intel i7-based workstation and camera \cite{kml_sensors}, two high-bandwidth microphones \cite{earthworks_sr314}, and an audio interface \cite{motu_m4}. The microphones were selected to enable studies on the observability of weld defects at higher frequencies. Moreover, audio samples were recorded using a 192 KHz sampling rate and saved in lossless FLAC format \cite{flac}. The video samples were recorded at a nominal 30 FPS and saved in AVI format. Fig.~\ref{fig:kml_camera_and_computer} shows the KML camera and computer. The camera was attached to the welder arm, about 200mm from the torch. The microphones were attached to the work table, about 300mm from the welder arm motion axis.

\begin{table}[ht]
    \centering
    \begin{tabular}{|c|c|c|}
    \hline
    \bf{Quantity} & \bf{Component} & \bf{Description}\\
    \hline
    1 & welding robot &  OTC AII-V6 6-axis arc welding robot\\
    \hline
    1 & camera  & KML welding camera\\
    \hline
    2 & microphone & Earthworks SR314\\
    \hline
    1 & microphone interface & MOTU M4 audio interface\\
    \hline
    1 & workstation & KML workstation with Intel i7 processor\\
    \hline
    \end{tabular}
\caption{\label{tab:workstation_table} Data collection station parts list.}
\end{table}

To generate weld defects, the supplier contracted a welding expert who supervised the initial configuration of the welding robot to produce the desired weld defects. The welding expert did not perform any post-weld validation or labeling. Thus, the dataset contains some amount of label noise. Moreover, all sensors were triggered to start recording at the same time, but there is some variation in the actual time it takes for the individual device to respond. To align the different modalities which is required for multi-modal experiments, the weld start and end times were identified for each modality by inspecting the audio and illumination changes at the start and end of welding. The audio data, which was collected through a dedicated audio interface with a low-latency driver \cite{motu_m4} was taken as the ground truth source for determining the welding duration. In this way, it was determined that the actual recorded frames-per-second (FPS) of the video varies from the expected 30 FPS. It is worth mentioning that this post-collection alignment would not be needed if the robot command signals were readily accessible.

\subsection{Welding Sample Distribution}
The dataset contains weld samples for the 12 weld categories shown in Tab.~\ref{tab:weld_categories}. Each weld category has been recorded for different weld types, and, where applicable, for different materials. Tab.~\ref{tab:samples_summary} contains a brief summary of the dataset by weld type. A complete breakdown of the dataset is included in  Appendix~\ref{secA1}. Fig.~\ref{fig:welding_samples_1} in Appendix~\ref{secA2} shows some examples of (post-weld) photos of welding samples.
\begin{table}[ht]
    \centering
    \begin{tabular}{|c|c|}
    \hline
        \bf{Weld Category} &  \bf{Total} \\
        \hline
        Good & 819 \\        
        \hline
        Excessive Convexity & 160\\
        \hline
        Undercut & 160\\
        \hline
        Crater Cracks & 161 \\
        \hline
        Overlap & 160\\
        \hline
        Excessive Penetration & 480\\
        \hline
        Porosity w/Excessive Penetration & 480 \\
        \hline
        Spatter & 320 \\
        \hline
        Lack Of Fusion & 320 \\
        \hline
        Warping & 320 \\
        \hline
        Porosity & 340\\
        \hline
        Burnthrough & 320\\
        \hline
        \bf{All} & 4040\\
        \hline
    \end{tabular}
\caption{\label{tab:weld_categories} Summary of the number of weld samples by weld category.} 
\end{table}

\begin{table}[ht]
    \centering
    \begin{tabular}{|c|c|c|c|c|}
    \hline
        \bf{Weld Type} & \bf{Samples} & \bf{Material} & \bf{Samples} & \bf{Total} \\
        \specialrule{.1em}{.05em}{.05em}
        non-fillet & 1720 & 7mm-FE410 & 2560 & 2919 \\
        \hline
        non-fillet & 840 & 3mm-FE410 & ~ & ~ \\
        \hline
        non-fillet & 239 & 7mm-BSK46 & 359 & ~ \\
        \hline
        non-fillet & 120 & 3mm-BSK46 & ~ & ~ \\
        \specialrule{.1em}{.05em}{.05em}
        fillet & 701 & 7mm-FE410 & 981 & 1121 \\
        \hline
        fillet & 280 & 3mm-FE410 & ~ & ~ \\ 
        \hline
        fillet & 100 & 7mm-BSK46 & 140 & ~ \\ 
        \hline
        fillet & 40 & 3mm-BSK46 & ~ & ~\\ 
        \specialrule{.1em}{.05em}{.05em}
    \end{tabular}
\caption{\label{tab:samples_summary} Summary of the number of weld samples by weld type and material used.} 
\end{table}

\section{Experimental Setup}
The experiments in this work focus on the following questions: (1) Can weld defects be reliably detected in real-time from audio and/or video recordings? (2) How does the detection accuracy depend on the defect type and modality? (3) What improvements can be expected from combining audio and video in a multi-modal system?

To answer those questions, we treat the weld defect problem as an anomaly detection problem, not a classification problem. That means, our models do not attempt to classify a sample into a weld category, but generate an anomaly score that increases with the likelihood of a defect. For a real use case a threshold has to be determined in advance. If the score of the model exceeds this threshold during welding, a defect will be detected and the supervisor of the robot will be notified. The choice of the threshold depends on the false positive and false negative rates of the model for the defect types that are relevant for a specific use case, the costs associated with a false detection of a defect, and the costs resulting from a missed defect. As we would like to evaluate the quality of our models independently of an application-specific threshold, we compare the models using the Area-under-ROC-Curve (AUC) metric which is scale- and threshold-independent \cite{fawcett_2006}. AUC is defined as the area under the Receiver Operating Characteristic (ROC) curve for all possible values of the false positive rate (FPR):
\begin{equation}
\textit{AUC} = \int_{x=0}^1 \textit{ROC}(x) dx
\end{equation}
where $x$ is the FPR and $\textit{ROC}(x)$ is the true positive rate (TPR) \cite{auc2}. 

When calculating FPR and TPR for a specific threshold on the validation and test sets, we have to take into account that we are targeting a real-time scenario, where the model produces an anomaly score at every time frame. On the other hand, each sample in our data set has just a single label denoting the weld category. For instance, a sample could be labeled as ``porosity'' but there is no indication at which time instance the defect occurs. Therefore, we aggregate all scores produced by the model for all frames of the sample by either taking their maximum or expected value. 

In this work we employ simple but effective convolutional neural networks (CNN) for weld defect detection. This allows us to focus more on gaining insights and less on conducting large hyper-parameter optimizations and architecture searches. For audio, we used a 1D CNN auto-encoder \cite{lopez_2021}. For video, we applied the relatively simple 3D CNN from \cite{slow_fast} provided by the MMAction2 library \cite{mmaction2_2020}\footnote{The  ``slowfast\_r101\_4x16x1\_256e\_kinetics400\_rgb\_20210218-d8b58813.pth'' checkpoint from v0.15.0 was used \cite{slow_fast_checkpoint}.}. All our work used the PyTorch tensor library for Python \cite{pytorch,python}.

For the experiments, the dataset is split into a training, validation, and test partition. The training data includes only good, i.e., normal-state, samples, while the defects are divided equally among the validation and test subsets. 
The validation partition of the dataset is used for hyperparameter tuning. The best hyperparameter configuration on the validation partition is used to generate the test results from the test partition. The data split is summarized in Tab.~\ref{tab:datasplit}. The same partitions were used for all experiments with both modalities.

\begin{table}[ht]
    \centering
    \begin{tabular}{|c|c|c|}
    \hline
        \bf{Partition} & \bf{Number of ``Good'' Samples} & \bf{Number of ``Defective'' Samples} \\
        \hline
        Train & 576 & 0  \\
        \hline
        Validation & 122 & 1610\\
        \hline
        Test & 121 & 1611 \\
        \hline
    \end{tabular}
\caption{\label{tab:datasplit} Data split into training, validation, and test partitions.} 
\end{table}

\subsection{Acoustic Anomaly Detection}
For the audio experiments, the original sampling rate of 192~kHz was maintained and only one channel was used. The audio CNN auto-encoder architecture is shown in Table \ref{tab:audio_cnn}. A detailed description of the model's topology can be found in \cite{lopez_2021}: the key characteristic is that the bottleneck layer largely preserves the time dimension. This feature has been motivated by the work of Agrawal et al. \cite{agrawal_2020}. The model in Tab.~\ref{tab:audio_cnn} is much smaller than the one described in \cite{agrawal_2020} because the convolutions are not gated.

\begin{table}[ht]
    \centering
    \begin{tabular}{|c|c|c|c|}
    \hline
        \bf{Layer} & \bf{Input, Output Channels} & \bf{Kernel Size} & \bf{Stride} \\
        \hline
        BatchNorm1D & n-bins & N/A & N/A\\
        \hline
        Conv1D & (n-bins, 1024) & 3 & 1\\
        \hline
        3 x Conv1D & (1024, 1024) & 3 & 1\\
        \hline
        Conv1D  & (1024, bottleneck size) & 3 & 1\\
        \hline
        ConvTranspose1D  & (bottleneck size, 1024) & 3 & 1\\
        \hline
        3 x ConvTranspose1D & (1024, 1024) & 3 & 1\\
        \hline
        ConvTranspose1D & (1024, n-bins) & 3 & 1\\
        \hline
    \end{tabular}
\caption{\label{tab:audio_cnn} Audio 1D CNN.} 
\end{table}
 The model uses leaky-ReLU activations in all but the last activation before the output layer, which uses a PReLU activation. Overall, the model has 31,670,306 trainable parameters.

 The latency of this model equals the hop length used to train the model. For example, for a hop length of 8192 and 192~kHz audio, the latency is about 42.7\,ms. The model has 5 encoding layers with kernel size 3. Each encoding layer decreases the time dimension by 2, therefore inputs must have more than 10 frames. This means the input buffer must satisfy Eq.~\ref{eq:buffer_size}, with the FFT window an integer multiple of the hop length. In the foregoing example, the input buffer needs to hold $12 \times 8192$ samples.

 \begin{equation}
     \text{buffer size} = \text{hop length} \times \left( 10 + \frac{\text{FFT window}}{\text{hop length}} \right)
     \label{eq:buffer_size}
 \end{equation}

\subsection{Visual Anomaly Detection}

Our approach to visual weld defect detection is based on a two-stage process. The first stage encodes each video frame into a fixed dimensional feature vector. For this, a window that is 64 frames long is shifted frame-by-frame over the whole video. 64 frames correspond to roughly 2 seconds at a frame rate of 30 FPS. As the window is centered around the frame of interest, the 64-frame window size leads to defect detection latency of around one second in a real-time scenario. This is a larger latency than for the acoustic anomaly detection, which is based on much smaller windows as will be described in Sec.~\ref{sec:results}. Still, we believe that it should be acceptable for many use cases. Each window is encoded into a $2304$-dimensional feature vector using the pre-trained and fixed Slowfast \cite{slow_fast} model. 
In the second stage, an auto-encoder model consisting of an encoder, a bottleneck, and a decoder is used for generating anomaly scores from the input feature vector. The encoder maps the $2304$-dimensional input feature vector to $64$-dimensional latent space by passing it through multiple linear layers along with ReLU activation and dropout in sequence. Each linear layer reduces the dimension by $\frac{1}{2}$, thus creating a bottleneck with a $64$-dimensional latent vector. Later, the decoder uses the latent space embedding to reconstruct the original feature of the frame. The decoder consists of multiple linear layers with ReLU activation and dropout as well. Each linear layer of the decoder scales the dimension by 2 until the original $2304$ dimension is obtained. All dropout layers zero weights with probability $0.5$.  The architecture details of the auto-encoder are provided in Tab.~\ref{tab:video_cnn}. 

The auto-encoder model is trained on welding videos minimizing the model's anomaly score which is defined as the mean-squared-error (MSE) between the input feature vector and the output of the decoder. Note, that the Slowfast model from the first stage is fixed, and no-back propagation is applied to the first stage during training.

\begin{table}[ht]
    \centering
    \begin{tabular}{|c|c|}
    \hline
        \bf{Layer} & \bf{Input, Output Dim.} \\
        \hline
        Linear & (2304, 512) \\
        \hline
        Linear & (512, 256) \\
        \hline
        Dropout ($p=0.5$) & - \\
        \hline
        Linear & (256, 128) \\
        \hline
        Dropout  ($p=0.5$) & - \\
        \hline
        Linear & (128, 64) \\
        \hline
        Dropout  ($p=0.5$) & - \\
        \hline
        Linear & (64, 64) \\
        \hline
        \hline
        Linear & (64, 64) \\
        \hline
        Linear & (64, 128) \\
        \hline
        Linear & (128,256) \\
        \hline
        Linear & (256, 512) \\
        \hline
        Dropout  ($p=0.5$) & - \\
        \hline
        Linear & (512, 2304) \\
        \hline
        
    \end{tabular}
\caption{\label{tab:video_cnn} Auto-encoder model for the video modality.} 
\end{table}

\subsection{Multi-modal Anomaly Detection} \label{multi_model_subsection}
We used a late-fusion approach to combine the anomaly scores of the two modalities. Since anomaly scores generated by different models generally have different scales, we first standardized the scores using the mean and standard deviation of the anomaly scores over the training set. Next, we identified the optimal convex combination of the audio and video anomaly scores by running grid search over convex combinations on the validation data. The best weighting is applied to compute the anomaly scores on the test set and to produce the final results.

\section{Results}
\label{sec:results}
\subsection{Acoustic Anomaly Detection}
For hyperparameter tuning we trained separate auto-encoder models with different parameter settings on the training partition and evaluated their AUC on the validation set. More specifically, a grid search was performed over FFT window sizes 4096, 16384, 32768, and 65536 and bottleneck dimensions 16, 32, 48, and 64. The hop length was fixed at 50\% of the FFT window size. We remind the reader that for a given analysis (time) window, the corresponding FFT window length is proportional to the sampling rate. Therefore, the FFT window sizes included in the grid search range from approximately 21\,ms to 341\,ms for the sampling rate in our data set, which is 192\,kHz.  The latency of these models range from approximately 11ms to 171ms. All models were trained for 50 epochs using a one-cycle learning schedule with a peak learning rate of $1\times 10^{-4}$ for the Adam optimizer \cite{smith_2018,adam_optimizer}. The MSE loss was minimized during training.

Tab.~\ref{tab:fft_window_search} shows the FFT window search results. These experiments determine that an FFT window of 16384 with a bottleneck dimension of 48 produced the best performance on the validation set. To obtain the AUC scores we explored several frame-wise anomaly score aggregation methods: expected value, moving average (MA) smoothing, and taking the max. We found taking the average worked best, followed closely by MA smoothing, and taking taking the max was last. We only show the expected value scores for space considerations.

\begin{table}[ht]
\centering
\begin{tabular}{|c|c|c|c|c|}
\hline
 \bf{FFT} &  \bf{Val AUC} &  \bf{Val AUC} &  \bf{Val AUC} &  \bf{Val AUC} \\
 \bf{Window} & \bf{(b.n.=16)} & \bf{(b.n.=32)}  & \bf{(b.n.=48)} & \bf{(b.n.=64)} \\
\hline
   4096    &    0.7550    &    0.7953    &    0.7597    &    0.7549\\
\hline
   16384    &    \bf{0.8444}   &    \bf{0.8434}   &    \bf{0.8451}    &    \bf{0.8426}\\
\hline
   32768    &    0.8210    &    0.8231    &    0.8262    &    0.8187\\
\hline
   65536    &    0.8193    &    0.8206    &    0.8215    &    0.8234\\
\hline
\end{tabular}
\caption{\label{tab:fft_window_search}FFT Window and Bottleneck Search.}
\end{table}

\begin{table}[ht]
\centering
\Large
\resizebox{\columnwidth}{!}{
\begin{tabular}{|c|c|c|c|c|}
\hline 
\bf{Welding Category}& \bf{FFT=4096} & \bf{FFT=16384} & \bf{FFT=32768} & \bf{FFT=65536} \\
\hline
Excessive Penetration & 0.7023 & 0.7717 & 0.8157 & \bf{0.8268} \\
\hline
Burnthrough & \bf{0.8256} & 0.6920 & 0.5952 & 0.6256 \\
\hline
Porosity & \bf{0.9898} & 0.9892 & 0.9633 & 0.9385\\
\hline
Porosity w/EP & 0.9584 & \bf{0.9758} & 0.9501 & 0.9460\\        
\hline
Undercut & \bf{0.9527} & 0.9143 & 0.8957 & 0.9000 \\
\hline
Overlap & 0.9800 & 0.9840 & 0.9863 & \bf{0.9874}\\
\hline 
Lack of fusion & 0.4702 & 0.8316 & \bf{0.8345} & 0.8226\\
\hline
Excessive Convexity & \bf{0.8981} & 0.8845 & 0.8704 & 0.8548\\
\hline 
Spatter & 0.3591 & 0.7773 & 0.8531 & \bf{0.8906} \\
\hline
Warping & 0.6411 & \bf{0.7518} & 0.5925 & 0.5030\\
\hline
Crater Cracks & \bf{0.9912} & 0.9672 & 0.9678 & 0.9659\\
\specialrule{.1em}{.05em}{.05em}
\bf{All} & 0.7597 & \bf{0.8451} & 0.8262 & 0.8215\\
\hline
\end{tabular}
}
\caption{\label{tab:fft_window_search_by_defect} Validation AUC by FFT window size and defect type for model with bottleneck=48.}
\end{table}

\begin{table}[ht]
\centering
\Large
\resizebox{\columnwidth}{!}{
\begin{tabular}{|c|c|c|c|c|}
\hline 
\bf{Welding Category}& \bf{FFT=4096} & \bf{FFT=16384} & \bf{FFT=32768} & \bf{FFT=65536} \\
\hline
Excessive Penetration & 0.7023 & 0.7807 & \bf{0.8220} & 0.8204\\
\hline
Burnthrough & \bf{0.8250} & 0.7018 & 0.6202 & 0.6497\\
\hline
Porosity & 0.9896 & \bf{0.9949} & 0.956 & 0.9144 \\
\hline
Porosity w/EP & \bf{0.9663} & 0.9659 & 0.9360 & 0.9155\\        
\hline
Undercut & \bf{0.9489} & 0.8989 & 0.8811 & 0.8637 \\
\hline
Overlap & 0.9695 & \bf{0.9743} & 0.9671 & 0.9631 \\
\hline 
Lack of fusion & 0.4693 & \bf{0.8372} & 0.8353 & 0.8211 \\
\hline
Excessive Convexity & \bf{0.8921} & 0.8842 & 0.8553 & 0.8361\\
\hline 
Spatter & 0.3855 & 0.7796 & 0.8407 & \bf{0.8678} \\
\hline
Warping & 0.6483 & \bf{0.7648} & 0.6723 & 0.6060 \\
\hline
Crater Cracks & \bf{0.9935} & 0.9453 & 0.9218 & 0.9093 \\
\specialrule{.1em}{.05em}{.05em}
\bf{All} & 0.7651 & \bf{0.8460} & 0.8296 & 0.8157\\
\hline
\end{tabular}
}
\caption{\label{tab:test_fft_window_search_by_defect} Test AUC by FFT window size and defect type for model with bottleneck=48.}
\end{table}

On the test set, the best validation model obtained an AUC of $0.8460$. This is not much different from the AUC on the validation set which indicates that the model does not overfit.
To gain insights into the difficulty of detecting particular defects, we also determined the validation and test AUC by defect type -- the results are shown in Tab.~\ref{tab:fft_window_search_by_defect} and Tab.~\ref{tab:test_fft_window_search_by_defect}, respectively. Again, the results indicate a high agreement between validation and test set AUCs. Note that these results are not directly actionable because in most use cases the defect type is not known a priori. Breaking-out performance by defect type is, however, useful for estimating performance gains that could be obtained when fusing audio with video information.

In cases where an error (or error cost) distribution is known a priori, one may select the optimal hyperparameters differently. Additionally, latency or memory requirements can affect the FFT window and hop length selection as well.

\subsection{Visual Anomaly Detection}

For visual defect detection, we first generate feature vectors of dimension $2304$ using the pre-trained Slowfast model for each sample in all data partitions. In a second step we train the auto-encoder model using the Adam optimizer with a learning rate of $0.0005$ and MSE loss function for up to $1000$ epochs on the feature vectors. For the model that performs best on the validation set, we compare different methods to aggregate the scores:
simply taking the maximum score (``Max w/o smoothing''), smoothing the scores by averaging them within a one-second window (``Max over 1s-MA''), or smoothing the scores by averaging them within the full two-second window (``Max over 2s-MA''). Table \ref{tab:video_results_test} shows the AUC for different aggregation methods on the test set. The best validation model obtains an overall AUC of $0.9052$ and $0.8977$ on the validation and test data, respectively.

\begin{table}[ht]
    \centering
   \begin{tabular}{|c|c|c|c|}
    \hline
         & \bf{Max w/o Smoothing} & \bf{Max Over 1s-MA} & \bf{Max Over 2s-MA} \\
        \bf{Weld Category} & \bf{(AUC)} & \bf{(AUC)} & \bf{(AUC)} \\
        \hline
        Excessive Penetration & 0.8165  & 0.8815  & 0.9011  \\
        \hline
        Burnthrough & 0.8066  & 0.8978  & 0.9156  \\
        \hline
        Porosity & 0.9732  & 0.9984  & 0.9998  \\
        \hline
        Porosity w/EP & 0.9701  & 0.9985  & 0.9999  \\        
        \hline
        Undercut & 0.9180  & 0.9462  & 0.9575  \\
        \hline
        Overlap & 0.7446  & 0.7973  & 0.8155  \\
        \hline 
        Lack of fusion & 0.7229  & 0.7461  & 0.7671  \\
        \hline
        Excessive Convexity & 0.9809  & 0.9965  & 0.9984  \\
        \hline 
        Spatter & 0.9458  & 0.9793  & 0.9857  \\
        \hline
        Warping & 0.6690  & 0.6991  & 0.7141  \\
        \hline
        Crater Cracks & 0.9020  & 0.8597  &  0.8635 \\
        \specialrule{.1em}{.05em}{.05em}
        \bf{All} & 0.8577  &  0.8947 &  \bf{0.9052} \\
        \hline
    \end{tabular}
\caption{\label{tab:video_results_val} Defect specific AUC for various anomaly score aggregation methods calculated for the validation data using video.} 
\end{table}

\begin{table}[ht]
    \centering
   \begin{tabular}{|c|c|c|c|}
    \hline
         & \bf{Max w/o Smoothing} & \bf{Max Over 1s-MA} & \bf{Max Over 2s-MA} \\
        \bf{Weld Category} & \bf{(AUC)} & \bf{(AUC)} & \bf{(AUC)} \\
        \hline
        Excessive Penetration & 0.7746 & 0.8534 & 0.8747 \\
        \hline
        Burnthrough & 0.7674 & 0.8841 & 0.8997 \\
        \hline
        Porosity & 0.9557 & 0.9847 & 0.9868 \\
        \hline
        Porosity w/EP & 0.9558 & 0.9846 & 0.9873 \\        
        \hline
        Undercut & 0.8744 & 0.9159 & 0.9270 \\
        \hline
        Overlap & 0.7530 & 0.8319 & 0.8516 \\
        \hline 
        Lack of fusion & 0.7055 & 0.7644 & 0.7784 \\
        \hline
        Excessive Convexity & 0.9673 & 0.9825 & 0.9844 \\
        \hline 
        Spatter & 0.9265 & 0.9583 & 0.9654 \\
        \hline
        Warping & 0.6632 & 0.7375 & 0.7456 \\
        \hline
        Crater Cracks & 0.8527 & 0.8323 & 0.8436 \\
        \specialrule{.1em}{.05em}{.05em}
        \bf{All} & 0.8345 & 0.8873 & \bf{0.8977} \\
        \hline
    \end{tabular}
\caption{\label{tab:video_results_test} Defect specific AUC for various anomaly score aggregation methods calculated for the test data using video.} 
\end{table}

\subsection{Multi-modal Anomaly Detection}
 The input scores for the late fusion combination of the video and audio modalities were created using the best performing models for each modality, i.e., the audio model using FFT window size 16384 and bottleneck dimension of 48, and the video model with 2s-MA smoothing. We determined the weighting for the modalities on the validation data using the grid search described in Sec.~\ref{multi_model_subsection} with step size $0.01$.

This way, we found the best weighting to be $0.37$ and $0.63$ for the audio and video scores, respectively. These weightings make sense given the stronger performance of the video modality.

Using the optimal weighting we obtained our overall best test AUC of $0.9178$. From Tab.~\ref{tab:multi_model_results}, it can be seen that the overall audio validation and test AUCs improved by 9.8\% and 8.5\%, respectively, and the overall video validation and test AUCs improved by 2.5\% and 2.2\%, respectively. The defect-specific scores  improved as well. For audio, the AUC metric improved for 8 of 11 defect categories. For video, the AUC improved for 7 of 11 categories. It is worth mentioning that the weakest performing defect categories all improved. In particular, audio had four defect categories with AUCs in the $0.7$s and all benefited from fusing. For video, two defect categories had AUCs in the $0.7$s and both benefited from fusing.  See the entries of Tab.~\ref{tab:multi_model_delta} shown in bold.

\begin{figure}[ht]
\centering
\includegraphics[scale=0.5]{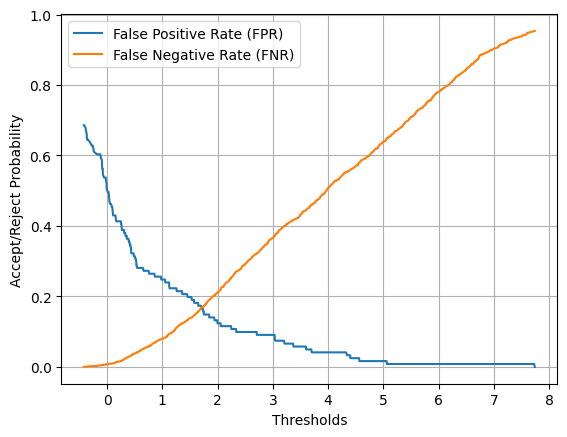}
\caption{Accept-reject probability of multi-modal predictions on test data.}\label{fig:multi_modal_test_det_curve}
\end{figure}

Fig.~\ref{fig:multi_modal_test_det_curve} shows the detection error curves for the multi-modal predictions, on the test data. As can be seen, the FPR and false-negative-rate (FNR) intersect at about 17\%.

\begin{table}[ht]
\centering
\begin{tabular}{|c|c|c|}
\hline
 \bf{Weld Category} &  \bf{Validation (AUC)} &  \bf{Test (AUC)} \\
\hline
Excessive Penetration & 0.9134 & 0.8893 \\
\hline
Burnthrough & 0.8811 & 0.8713 \\
\hline
Porosity & 1 & 0.9901 \\
\hline
Porosity w/EP & 0.9997 & 0.9884 \\        
\hline
Undercut & 0.9800 & 0.9477 \\
\hline
Overlap & 0.9573 & 0.9310 \\
\hline 
Lack of fusion & 0.8306 & 0.8455 \\
\hline
Excessive Convexity & 0.9982 & 0.9823 \\
\hline 
Spatter & 0.9809 & 0.9576 \\
\hline
Warping & 0.7671 & 0.7984 \\
\hline
Crater Cracks & 0.9567 & 0.9266 \\
\specialrule{.1em}{.05em}{.05em}
\textbf{All} & 0.9280 & 0.9178 \\ 
\hline
\end{tabular}
\caption{\label{tab:multi_model_results} Multi-modal performance on validation and test data.}
\end{table}

\begin{table}[ht]
\centering
\begin{tabular}{|c|c|c|}
\hline
 \bf{Weld Category} &  \bf{Audio Change \%} &  \bf{Video Change \%} \\
\hline
Excessive Penetration & \bf{13.9215} & 1.6732 \\
\hline
Burnthrough & \bf{24.1646} & -3.1575 \\
\hline
Porosity & 3.9451 & 0.3350 \\
\hline
Porosity w/EP & 5.7199 & 0.1151 \\        
\hline
Undercut & 5.5453 & 2.2401 \\
\hline
Overlap & -4.2601 & 9.3291 \\
\hline 
Lack of fusion & 0.9933 & \bf{8.6198} \\
\hline
Excessive Convexity & 11.1384 & -0.2099 \\
\hline 
Spatter & \bf{22.8806} & -0.8132 \\
\hline
Warping & \bf{4.9216} & \bf{7.0805} \\
\hline
Crater Cracks & -1.6248 & 9.8452 \\
\specialrule{.1em}{.05em}{.05em}
\textbf{All} & 8.4811  & 2.2337 \\ 
\hline
\end{tabular}
\caption{\label{tab:multi_model_delta} Multi-modal test data performance difference by modality and defect. The bolded entries show the changes on defects with unimodal AUCs in the 0.7s.}
\end{table}

\section{Conclusions and Future Work}
In this work, we explored unsupervised weld defect detection using audio, video, and their combination. We demonstrated that, using a deep-learning based approach, both modalities allow to reliably detect the most important defect types in real-time. The best audio model had a latency of 42.7ms and is best suited for detecting porosity, overlap, and crater cracks. It has the lowest AUC scores for burnthrough, excessive penetration, spatter, and warping. Video generally shows better AUC than audio, but this comes at the price of a larger overall model size and a higher latency. Using video, the best detection performance is achieved for porosity, excessive convexity, and spatter. We observe the lowest performance for lack of fusion, and warping. We demonstrated that a combined approach using late fusion of normalized scores for both modalities offers improvements. More specifically, AUC scores for all defect types average 0.92, with the lowest score approximately 0.80. The worst performing categories for both modalities all improved. 

Future work will investigate more elaborate ways to combine the two modalities, e.g., a joint model that directly incorporates input from all sensors. Furthermore, we believe that the biggest limitation of our dataset is that it has been collected in a supervised way, i.e., the robot has been configured by purpose to generate defects as we required a sufficient number of defect samples for our experiments. This approach creates a potential mismatch to a real use case where defects occur randomly and rarely. We expect that defects will show up  less pronounced than in our dataset. Therefore, we plan to record a real production process of a collaboration partner and to label the defects as they occur. This will allow us to validate our proposed methods in a more realistic setting.

\section*{Compliance with Ethical Standards}
This research was funded by the Intel Corporation. We are unaware of any conflicts of interest. The welding data used in this work does not contain information that could be used to identify an individual. All work was done in a manner consistent with expected ethical standards.

\begin{appendices}

\section{Sample Distributions}\label{secA1}

This appendix contains a complete dataset description by defect category: Tab.~\ref{tab:good_table} shows the distribution of the samples for the normal-state, i.e., good weld category, by weld type and material. Tab.~\ref{tab:excessive_convexity_table} to \ref{tab:burnthrough_table} show the sample distribution for the remaining weld categories.

\begin{table}[ht]
    \centering
    \resizebox{\columnwidth}{!}{%
    \begin{tabular}{|c|c|c|c|c|c|}
    \hline
        \bf{Weld Category} & \bf{Weld Type} & \bf{Samples} & \bf{Material} & \bf{Samples} & \bf{Total} \\
        \hline
        Good & non-fillet & 580 & 7mm-FE410 & 580 & 659 \\ 
        \hline
        Good & non-fillet & 0 & 3mm-FE410 & ~ & ~ \\ 
        \hline
        Good & non-fillet & 79 & 7mm-BSK46 & 79 & ~ \\ 
        \hline
        Good & non-fillet & 0 & 3mm-BSK46 & ~ & ~ \\ 
        \hline
        Good & fillet & 140 & 7mm-FE410 & 140 & 160 \\ 
        \hline
        Good & fillet & 0 & 3mm-FE410 & ~ & ~ \\ 
        \hline
        Good & fillet & 20 & 7mm-BSK46 & 20 & ~ \\ 
        \hline
        Good & fillet & 0 & 3mm-BSK46 & ~ & ~ \\ 
        \hline
    \end{tabular}
    }
\caption{\label{tab:good_table} The normal-state welding sample distribution.}    
\end{table}

\begin{table}[ht]
    \centering
    \resizebox{\columnwidth}{!}{%
    \begin{tabular}{|c|c|c|c|c|c|}
    \hline
        \bf{Weld Category} & \bf{Weld Type} & \bf{Samples} & \bf{Material} & \bf{Samples} & \bf{Total} \\
        \hline
        Excessive Convexity & non-fillet & 0 & 7mm-FE410 & 0 & 0 \\
        \hline
        Excessive Convexity & non-fillet & 0 & 3mm-FE410 & ~ & ~ \\
        \hline
        Excessive Convexity & non-fillet & 0 & 7mm-BSK46 & 0 & ~ \\
        \hline
        Excessive Convexity & non-fillet & 0 & 3mm-BSK46 & ~ & ~ \\
        \hline
        Excessive Convexity & fillet & 140 & 7mm-FE410 & 140 & 160 \\
        \hline
        Excessive Convexity & fillet & 0 & 3mm-FE410 & ~ & ~ \\
        \hline
        Excessive Convexity & fillet & 20 & 7mm-BSK46 & 20 & ~ \\
        \hline
        Excessive Convexity & fillet & 0 & 3mm-BSK46 & ~ & ~\\
        \hline        
    \end{tabular}
    }
\caption{\label{tab:excessive_convexity_table} The distribution of welding samples containing excessive convexity.}    
\end{table}
       
\begin{table}[ht]
    \centering
    \resizebox{\columnwidth}{!}{%
    \begin{tabular}{|c|c|c|c|c|c|}
    \hline
        \bf{Weld Category} & \bf{Weld Type} & \bf{Samples} & \bf{Material} & \bf{Samples} & \bf{Total} \\
        \hline
        Undercut & non-fillet & 0 & 7mm-FE410 & 0 & 0 \\
        \hline
        Undercut & non-fillet & 0 & 3mm-FE410 & ~ & ~ \\
        \hline
        Undercut & non-fillet & 0 & 7mm-BSK46 & 0 & ~ \\
        \hline
        Undercut & non-fillet & 0 & 3mm-BSK46 & ~ & ~ \\
        \hline
        Undercut & fillet & 140 & 7mm-FE410 & 140 & 160 \\
        \hline
        Undercut & fillet & 0 & 3mm-FE410 & ~ & ~ \\
        \hline
        Undercut & fillet & 20 & 7mm-BSK46 & 20 & ~ \\
        \hline
        Undercut & fillet & 0 & 3mm-BSK46 & ~ & ~\\
        \hline        
    \end{tabular}
    }
\caption{\label{tab:undercut_table} The distribution of welding samples containing undercut.}    
\end{table}

\begin{table}[ht]
    \centering
    \resizebox{\columnwidth}{!}{%
    \begin{tabular}{|c|c|c|c|c|c|}
    \hline
        \bf{Weld Category} & \bf{Weld Type} & \bf{Samples} & \bf{Material} & \bf{Samples} & \bf{Total} \\
        \hline
        Crater Cracks & non-fillet & 0 & 7mm-FE410 & 0 & 0 \\
        \hline
        Crater Cracks & non-fillet & 0 & 3mm-FE410 & ~ & ~ \\
        \hline
        Crater Cracks & non-fillet & 0 & 7mm-BSK46 & 0 & ~ \\
        \hline
        Crater Cracks & non-fillet & 0 & 3mm-BSK46 & ~ & ~ \\
        \hline
        Crater Cracks & fillet & 141 & 7mm-FE410 & 141 & 161 \\
        \hline
        Crater Cracks & fillet & 0 & 3mm-FE410 & ~ & ~ \\
        \hline
        Crater Cracks & fillet & 20 & 7mm-BSK46 & 20 & ~ \\
        \hline
        Crater Cracks & fillet & 0 & 3mm-BSK46 & ~ &\\
        \hline        
    \end{tabular}
    }
\caption{\label{tab:crater_cracks_table} The distribution of welding samples containing crater cracks.}    
\end{table}

\begin{table}[ht]
    \centering
    \resizebox{\columnwidth}{!}{%
    \begin{tabular}{|c|c|c|c|c|c|}
    \hline
        \bf{Weld Category} & \bf{Weld Type} & \bf{Samples} & \bf{Material} & \bf{Samples} & \bf{Total} \\
        \hline
        Overlap & non-fillet & 0 & 7mm-FE410 & 0 & 0 \\
        \hline
        Overlap & non-fillet & 0 & 3mm-FE410 & ~ & ~ \\
        \hline
        Overlap & non-fillet & 0 & 7mm-BSK46 & 0 & ~ \\
        \hline
        Overlap & non-fillet & 0 & 3mm-BSK46 & ~ & ~ \\
        \hline
        Overlap & fillet & 140 & 7mm-FE410 & 140 & 160 \\
        \hline
        Overlap & fillet & 0 & 3mm-FE410 & ~ & ~ \\
        \hline
        Overlap & fillet & 20 & 7mm-BSK46 & 20 & ~ \\
        \hline
        Overlap & fillet & 0 & 3mm-BSK46 & ~ &\\
        \hline        
    \end{tabular}
    }
\caption{\label{tab:overlap_table} The distribution of welding samples containing overlap.}

\end{table}
\begin{table}[ht]
    \centering
    \resizebox{\columnwidth}{!}{%
    \begin{tabular}{|c|c|c|c|c|c|}
    \hline
        \bf{Weld Category} & \bf{Weld Type} & \bf{Samples} & \bf{Material} & \bf{Samples} & \bf{Total} \\
        \hline
        Excessive Penetration & non-fillet & 0 & 7mm-FE410 & 280 & 320 \\
        \hline
        Excessive Penetration & non-fillet & 280 & 3mm-FE410 & ~ & ~ \\
        \hline
        Excessive Penetration & non-fillet & 0 & 7mm-BSK46 & 40 & ~ \\
        \hline
        Excessive Penetration & non-fillet & 40 & 3mm-BSK46 & ~ & ~ \\
        \hline
        Excessive Penetration & fillet & 0 & 7mm-FE410 & 140 & 160 \\
        \hline
        Excessive Penetration & fillet & 140 & 3mm-FE410 & ~ & ~ \\
        \hline
        Excessive Penetration & fillet & 0 & 7mm-BSK46 & 20 & ~ \\
        \hline
        Excessive Penetration & fillet & 20 & 3mm-BSK46 & ~ &\\ 
        \hline        
    \end{tabular}
    }
\caption{\label{tab:excessive_penetration_table} The distribution of welding samples containing excessive penetration.}    
\end{table}

\begin{table}[ht]
    \centering
    \resizebox{\columnwidth}{!}{%
    \begin{tabular}{|c|c|c|c|c|c|}
    \hline
        \bf{Weld Category} & \bf{Weld Type} & \bf{Samples} & \bf{Material} & \bf{Samples} & \bf{Total} \\
        \hline
        Porosity w/ Excessive Penetration & non-fillet & 0 & 7mm-FE410 & 280 & 320 \\
        \hline
        Porosity w/ Excessive Penetration & non-fillet & 280 & 3mm-FE410 & ~ & ~ \\
        \hline
        Porosity w/ Excessive Penetration & non-fillet & 0 & 7mm-BSK46 & 40 & ~ \\
        \hline
        Porosity w/ Excessive Penetration & non-fillet & 40 & 3mm-BSK46 & ~ & ~ \\
        \hline
        Porosity w/ Excessive Penetration & fillet & 0 & 7mm-FE410 & 140 & 160 \\
        \hline
        Porosity w/ Excessive Penetration & fillet & 140 & 3mm-FE410 & ~ & ~ \\
        \hline
        Porosity w/ Excessive Penetration & fillet & 0 & 7mm-BSK46 & 20 & ~ \\
        \hline
        Porosity w/ Excessive Penetration & fillet & 20 & 3mm-BSK46 & ~ &\\
        \hline        
    \end{tabular}
    }
\caption{\label{tab:porosity_w_ep_table} The distribution of welding samples containing porosity and excessive penetration.}    
\end{table}

\begin{table}[ht]
    \centering
    \resizebox{\columnwidth}{!}{%
    \begin{tabular}{|c|c|c|c|c|c|}
    \hline
        \bf{Weld Category} & \bf{Weld Type} & \bf{Samples} & \bf{Material} & \bf{Samples} & \bf{Total} \\
        \hline
        Spatter & non-fillet & 280 & 7mm-FE410 & 280 & 320 \\
        \hline
        Spatter & non-fillet & 0 & 3mm-FE410 & ~ & ~ \\
        \hline
        Spatter & non-fillet & 40 & 7mm-BSK46 & 40 & ~ \\
        \hline
        Spatter & non-fillet & 0 & 3mm-BSK46 & ~ & ~ \\
        \hline
        Spatter & fillet & 0 & 7mm-FE410 & 0 & 0 \\
        \hline
        Spatter & fillet & 0 & 3mm-FE410 & ~ & ~ \\
        \hline
        Spatter & fillet & 0 & 7mm-BSK46 & 0 & ~ \\
        \hline
        Spatter & fillet & 0 & 3mm-BSK46 & ~ & \\
        \hline        
    \end{tabular}
    }
\caption{\label{tab:spatter_table} The distribution of welding samples containing spatter.}    
\end{table}

\begin{table}[ht]
    \centering
    \resizebox{\columnwidth}{!}{%
    \begin{tabular}{|c|c|c|c|c|c|}
    \hline
        \bf{Weld Category} & \bf{Weld Type} & \bf{Samples} & \bf{Material} & \bf{Samples} & \bf{Total} \\
        \hline
        Lack Of Fusion & non-fillet & 280 & 7mm-FE410 & 280 & 320 \\
        \hline
        Lack Of Fusion & non-fillet & 0 & 3mm-FE410 & ~ & ~ \\
        \hline
        Lack Of Fusion & non-fillet & 40 & 7mm-BSK46 & 40 & ~ \\
        \hline
        Lack Of Fusion & non-fillet & 0 & 3mm-BSK46 & ~ & ~ \\
        \hline
        Lack Of Fusion & fillet & 0 & 7mm-FE410 & 0 & 0 \\
        \hline
        Lack Of Fusion & fillet & 0 & 3mm-FE410 & ~ & ~ \\
        \hline
        Lack Of Fusion & fillet & 0 & 7mm-BSK46 & 0 & ~ \\
        \hline
        Lack Of Fusion & fillet & 0 & 3mm-BSK46 & ~ &\\
        \hline        
    \end{tabular}
    }
\caption{\label{tab:lack_of_fusion_table} The distribution of welding samples with insufficient fusion.}    
\end{table}

\begin{table}[ht]
    \centering
    \resizebox{\columnwidth}{!}{%
    \begin{tabular}{|c|c|c|c|c|c|}
    \hline
        \bf{Weld Category} & \bf{Weld Type} & \bf{Samples} & \bf{Material} & \bf{Samples} & \bf{Total} \\
        \hline
        Warping & non-fillet & 280 & 7mm-FE410 & 280 & 320 \\
        \hline
        Warping & non-fillet & 0 & 3mm-FE410 & ~ & ~ \\
        \hline
        Warping & non-fillet & 40 & 7mm-BSK46 & 40 & ~ \\
        \hline
        Warping & non-fillet & 0 & 3mm-BSK46 & ~ & ~ \\
        \hline
        Warping & fillet & 0 & 7mm-FE410 & 0 & 0 \\
        \hline
        Warping & fillet & 0 & 3mm-FE410 & ~ & ~ \\ 
        \hline
        Warping & fillet & 0 & 7mm-BSK46 & 0 & ~ \\
        \hline
        Warping & fillet & 0 & 3mm-BSK46 & ~ &\\
        \hline        
    \end{tabular}
    }
\caption{\label{tab:warping_table} The distribution of welding samples containing warping.}    
\end{table}

\begin{table}[ht]
    \centering
    \resizebox{\columnwidth}{!}{%
    \begin{tabular}{|c|c|c|c|c|c|}
    \hline
        \bf{Weld Category} & \bf{Weld Type} & \bf{Samples} & \bf{Material} & \bf{Samples} & \bf{Total} \\
        \hline
        Porosity & non-fillet & 300 & 7mm-FE410 & 300 & 340 \\
        \hline
        Porosity & non-fillet & 0 & 3mm-FE410 & ~ & ~ \\
        \hline
        Porosity & non-fillet & 40 & 7mm-BSK46 & 40 & ~ \\
        \hline
        Porosity & non-fillet & 0 & 3mm-BSK46 & ~ & ~ \\
        \hline
        Porosity & fillet & 0 & 7mm-FE410 & 0 & 0 \\
        \hline
        Porosity & fillet & 0 & 3mm-FE410 & ~ & ~ \\
        \hline
        Porosity & fillet & 0 & 7mm-BSK46 & 0 & ~ \\
        \hline
        Porosity & fillet & 0 & 3mm-BSK46 & &\\
        \hline        
    \end{tabular}
    }
\caption{\label{tab:porosity_table} The distribution of welding samples containing porosity.}    
\end{table}

\begin{table}[ht]
    \centering
    \resizebox{\columnwidth}{!}{%
    \begin{tabular}{|c|c|c|c|c|c|}
    \hline
        \bf{Weld Category} & \bf{Weld Type} & \bf{Samples} & \bf{Material} & \bf{Samples} & \bf{Total} \\
        \hline
        Burnthrough & non-fillet & 0 & 7mm-FE410 & 280 & 320 \\
        \hline
        Burnthrough & non-fillet & 280 & 3mm-FE410 & ~ & ~ \\
        \hline
        Burnthrough & non-fillet & 0 & 7mm-BSK46 & 40 & ~ \\
        \hline
        Burnthrough & non-fillet & 40 & 3mm-BSK46 & ~ & ~ \\
        \hline
        Burnthrough & fillet & 0 & 7mm-FE410 & 0 & 0 \\
        \hline
        Burnthrough & fillet & 0 & 3mm-FE410 & ~ & ~ \\
        \hline
        Burnthrough & fillet & 0 & 7mm-BSK46 & 0 & ~ \\
        \hline
        Burnthrough & fillet & 0 & 3mm-BSK46 \\
        \hline        
    \end{tabular}
    }
\caption{\label{tab:burnthrough_table} The distribution of welding samples containing burnthrough.}    
\end{table}

\section{Sample Images}\label{secA2}
\begin{figure}[ht]
     \centering
     \begin{subfigure}[b]{\textwidth}
         \centering
         \includegraphics[width=\textwidth]{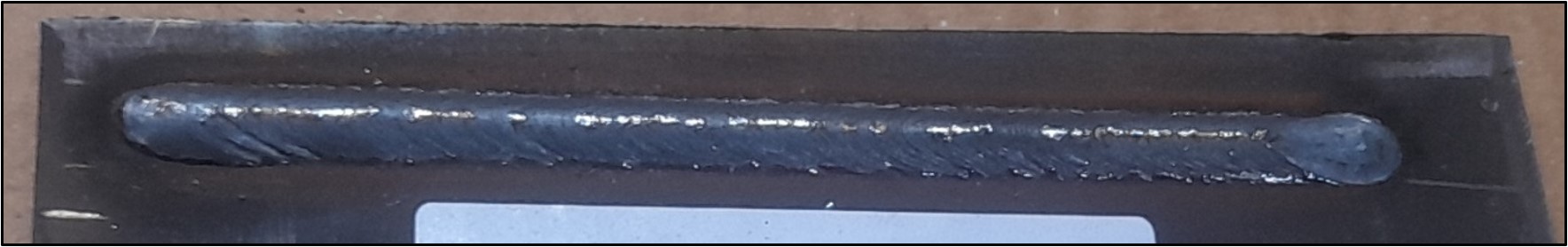}
         \caption{Good, Normal-condition, Sample.}
         \label{fig:good_sample}
     \end{subfigure}
     \vfill
     \begin{subfigure}[b]{\textwidth}
         \centering
         \includegraphics[width=\textwidth]{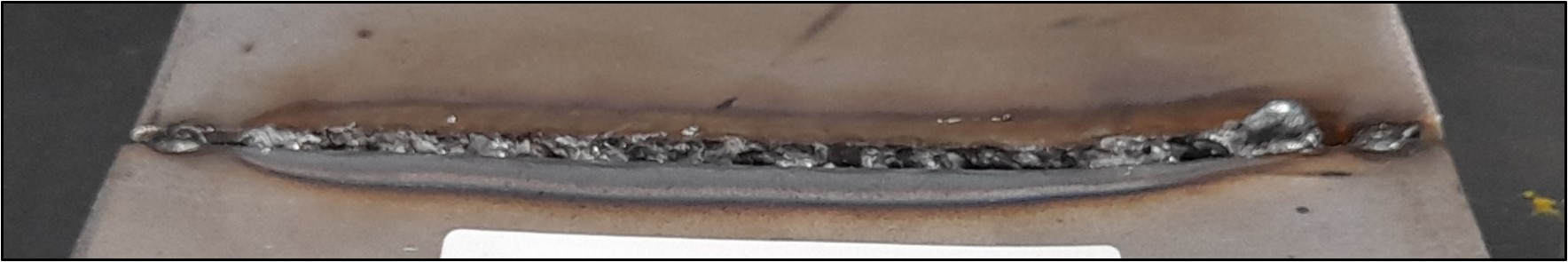}
         \caption{Burnthrough Sample.}
         \label{fig:burnthrough_sample}
     \end{subfigure}
     \vfill
     \begin{subfigure}[b]{\textwidth}
         \centering
         \includegraphics[width=\textwidth]{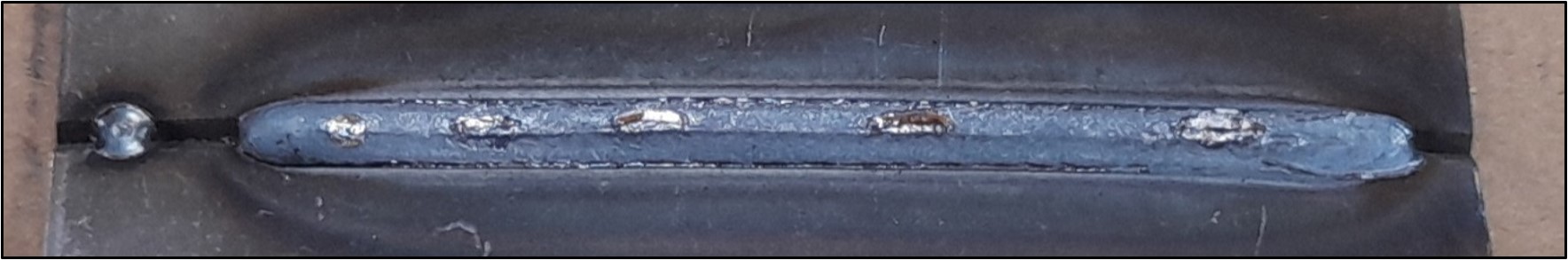}
         \caption{Excessive Penetration Sample.}
         \label{fig:excessive_penetration_sample}
     \end{subfigure}
	 \vfill
     \begin{subfigure}[b]{\textwidth}
         \centering
         \includegraphics[width=\textwidth]{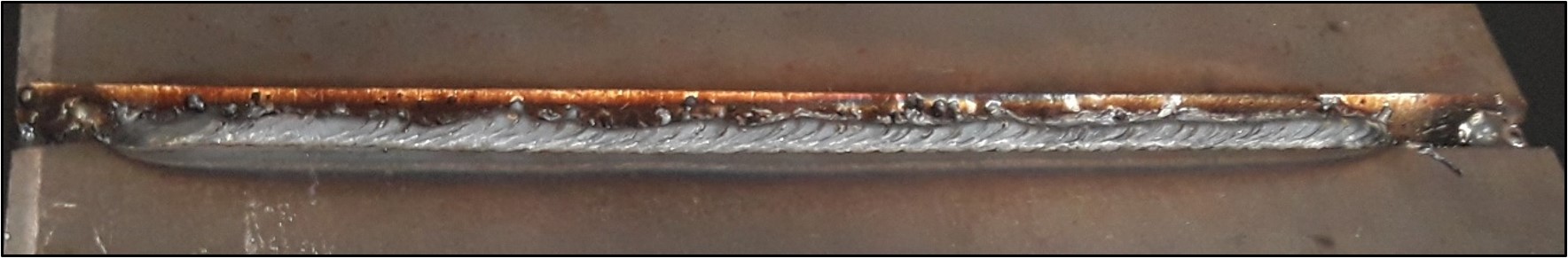}
         \caption{Lack Of Fusion Sample.}
         \label{fig:lack_of_fusion_sample}
     \end{subfigure}
	 \vfill
     \begin{subfigure}[b]{\textwidth}
         \centering
         \includegraphics[width=\textwidth]{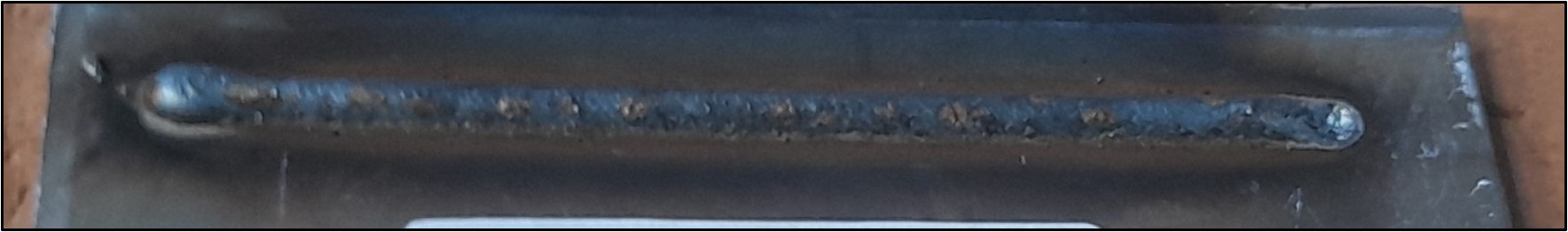}
         \caption{Porosity Sample.}
         \label{fig:porosity_sample}
     \end{subfigure}
	 \vfill
     \begin{subfigure}[b]{\textwidth}
         \centering
         \includegraphics[width=\textwidth]{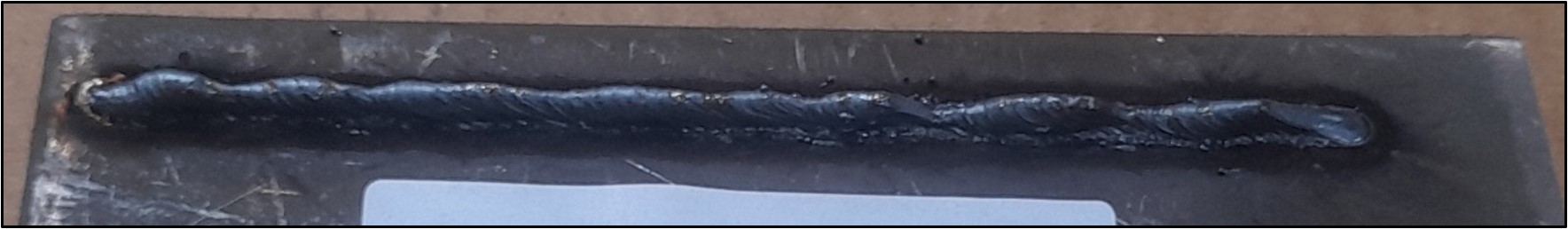}
         \caption{Spatter Sample.}
         \label{fig:spatter_sample}
     \end{subfigure}
	 \vfill
     \begin{subfigure}[b]{\textwidth}
         \centering
         \includegraphics[width=\textwidth]{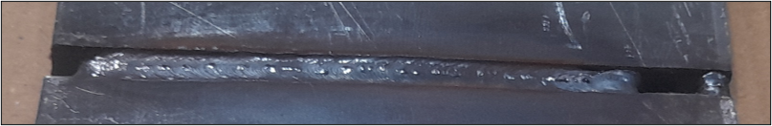}
         \caption{Warping Sample.}
         \label{fig:warping_sample}
     \end{subfigure}     
        \caption{Welding Samples}
        \label{fig:welding_samples_1}
\end{figure}

This Appendix shows in Fig.~\ref{fig:welding_samples_1} examples of photos taken of samples after the welding has been completed.

\end{appendices}


\bibliography{sn-bibliography}

\begin{thebibliography}{10}
\providecommand{\url}[1]{{#1}}
\providecommand{\urlprefix}{URL }
\providecommand{\doi}[1]{\url{https://doi.org/#1}}
\bibcommenthead

\bibitem{yu_2023}
N.~Lv, Y.~Xu, S.~Li, X.~Yu, S.~Chen, Deep learning based real-time and in-situ
  monitoring of weld penetration: Where we are and what are needed
  revolutionary solutions?
\newblock Journal of Manufacturing Processes \textbf{93}, 15--46 (2023).
\newblock \doi{https://doi.org/10.1016/j.jmapro.2023.03.011}

\bibitem{ma_2021}
G.~Ma, H.~Yuan, L.~Yu, Y.~He, Monitoring of weld defects of visual sensing
  assisted gmaw process with galvanized steel.
\newblock Materials and Manufacturing Processes \textbf{36}(10), 1178--1188
  (2021).
\newblock \doi{10.1080/10426914.2021.1885711}

\bibitem{zou_2020}
S.~Zou, Z.~Wang, S.~Hu, W.~Wang, Y.~Cao, Control of weld penetration depth
  using relative fluctuation coefficient as feedback.
\newblock Journal of Intelligent Manufacturing \textbf{31}(5), 1203--1213
  (2020).
\newblock \doi{10.1007/s10845-019-01506-8}

\bibitem{ji_2023}
T.~Ji, N.~Mohamad~Nor, Deep learning-empowered digital twin using acoustic
  signal for welding quality inspection.
\newblock Sensors \textbf{23}(5) (2023).
\newblock \doi{10.3390/s23052643}

\bibitem{alcaraz_2023}
J.Y.I. Alcaraz, W.~Foqu{\'e}, A.~Sharma, T.~Tjahjowidodo, Indirect porosity
  detection and root-cause identification in waam.
\newblock Journal of Intelligent Manufacturing  (2023).
\newblock \doi{10.1007/s10845-023-02128-x}

\bibitem{chen_2023}
L.~Chen, X.~Yao, C.~Tan, W.~He, J.~Su, F.~Weng, Y.~Chew, N.P.H. Ng, S.K. Moon,
  In-situ crack and keyhole pore detection in laser directed energy deposition
  through acoustic signal and deep learning.
\newblock Additive Manufacturing \textbf{69}, 103547 (2023).
\newblock \doi{10.1016/j.addma.2023.103547}

\bibitem{Madhvacharyula_2022}
A.~Madhvacharyula, A.~Pavan, S.~Gorthi, S.~Chitral, N.~Venkaiah, D.~Kiran, In
  situ detection of welding defects: a review.
\newblock Welding in the World \textbf{66} (2022).
\newblock \doi{10.1007/s40194-021-01229-6}

\bibitem{na_2021}
L.~Na, S.~jie Chen, Q.~heng Chen, W.~Tao, H.~Zhao, S.~ben Chen, Dynamic welding
  process monitoring based on microphone array technology.
\newblock Journal of Manufacturing Processes \textbf{64}, 481--492 (2021).
\newblock \doi{https://doi.org/10.1016/j.jmapro.2020.12.023}

\bibitem{lv_2017}
N.~Lv, Y.~Xu, S.~Li, X.~Yu, S.~Chen, Automated control of welding penetration
  based on audio sensing technology.
\newblock Journal of Materials Processing Technology \textbf{250}, 81--98
  (2017).
\newblock \doi{https://doi.org/10.1016/j.jmatprotec.2017.07.005}.
\newblock
  \urlprefix\url{https://www.sciencedirect.com/science/article/pii/S0924013617302777}

\bibitem{sumesh_2015}
A.~Sumesh, K.~Rameshkumar, K.~Mohandas, R.S. Babu, Use of machine learning
  algorithms for weld quality monitoring using acoustic signature.
\newblock Procedia Computer Science \textbf{50}, 316--322 (2015).
\newblock \doi{https://doi.org/10.1016/j.procs.2015.04.042}.
\newblock Big Data, Cloud and Computing Challenges

\bibitem{realitycheck_ad}
RealityCheck AD\\
  \url{https://www.renesas.com/us/en/products/microcontrollers-microprocessors/reality-ai/realitycheck-ad}
  (Last viewed March 27, 2023.)

\bibitem{realitycheck_ad_presentation}
RealityCheck AD Presentation\\
  \url{https://www.youtube.com/watch?v=-6B_XsEN2Q4} (Last viewed March 27,
  2023.)

\bibitem{amazon_monitron}
Amazon Monitron\\ \url{https://aws.amazon.com/monitron} (Last viewed March 27,
  2023.)

\bibitem{xiris_ai_product}
Xiris Audio AI\\
  \url{https://blog.xiris.com/blog/using-sound-and-imaging-for-detecting-welding-defects}
  (Last viewed March 27, 2023.)

\bibitem{eren_2023}
B.~Eren, M.~Demir, S.~Mistikoglu, Recent developments in computer vision and
  artificial intelligence aided intelligent robotic welding applications.
\newblock The International Journal of Advanced Manufacturing Technology
  (2023).
\newblock \doi{10.1007/s00170-023-11456-4}

\bibitem{breitenbach_2021}
J.~Breitenbach, T.~Dauser, H.~Illenberger, M.~Traub, R.~Buettner, in \emph{2021
  IEEE International Conference on Big Data (Big Data)} (2021), pp. 2019--2025.
\newblock \doi{10.1109/BigData52589.2021.9671887}

\bibitem{meyer_2022}
K.~Meyer, V.~Mahalec, Anomaly detection methods for infrequent failures in
  resistive steel welding.
\newblock Journal of Manufacturing Processes \textbf{75}, 497--513 (2022).
\newblock \doi{https://doi.org/10.1016/j.jmapro.2021.12.003}

\bibitem{asif_2022}
K.~Asif, L.~Zhang, S.~Derrible, E.~Indacochea, D.~Ozevin, B.~Ziebart, Machine
  learning model to predict welding quality using air-coupled acoustic emission
  and weld inputs.
\newblock Journal of Intelligent Manufacturing \textbf{33}, 1--15 (2022).
\newblock \doi{10.1007/s10845-020-01667-x}

\bibitem{mohanasundari_2021}
M.~L, K.~Senthilkumar, S.~Poruran, Performance analysis of weld image
  classification using modified resnet cnn architecture.
\newblock Turkish Journal of Computer and Mathematics Education (TURCOMAT)
  \textbf{12}, 2260--2266 (2021)

\bibitem{buongiorno_2022}
D.~Buongiorno, M.~Prunella, S.~Grossi, S.M. Hussain, A.~Rennola, N.~Longo,
  G.~Di~Stefano, V.~Bevilacqua, A.~Brunetti, Inline defective laser weld
  identification by processing thermal image sequences with machine and deep
  learning techniques.
\newblock Applied Sciences \textbf{12}(13) (2022).
\newblock \doi{10.3390/app12136455}.
\newblock \urlprefix\url{https://www.mdpi.com/2076-3417/12/13/6455}

\bibitem{graney_and_starry_2012}
B.~Graney, K.~Starry, Rolling element bearing analysis.
\newblock Materials evaluation \textbf{70} (2012)

\bibitem{Wang_2020}
B.~Wang, S.~Hu, L.~Sun, T.~Freiheit, Intelligent welding system technologies:
  State-of-the-art review and perspectives.
\newblock Journal of Manufacturing Systems \textbf{56} (2020).
\newblock \doi{10.1016/j.jmsy.2020.06.020}

\bibitem{motu_m4}
MOTU M4\\ \url{https://motu.com/en-us/products/m-series/m4} (Last viewed August
  3, 2023.)

\bibitem{auc2}
\url{https://en.wikipedia.org/wiki/Partial_Area_Under_the_ROC_Curve}(Last
  viewed December 7th, 2023.)

\bibitem{earthworks_sr314}
Earthworks SR314\\ \url{https://earthworksaudio.com/vocal-microphones/sr314}
  (Last viewed August 3, 2023.)

\bibitem{flac}
\url{https://xiph.org/flac/features.html}(Last viewed November 29, 2023.)

\bibitem{auc}
\url{https://en.wikipedia.org/wiki/Partial\_Area\_Under\_the\_ROC\_Curve}(Last
  viewed November 29, 2023.)

\bibitem{otc_v6}
OTC AII V6\\ \url{http://www.micharc.com/pdfs/otc/OTC%20AII%20Arc%20Robots.pdf}
  (Last viewed August 3, 2023.)

\bibitem{kml_sensors}
KML Sensors\\ \url{https://www.kmlsensors.com} (Last viewed August 3, 2023.)

\bibitem{slow_fast_checkpoint}
Slowfast checkpoint\\
  \url{https://download.openmmlab.com/mmaction/recognition/slowfast/slowfast\_r101\_4x16x1\_256e\_kinetics400\_rgb/slowfast\_r101\_4x16x1\_256e\_kinetics400\_rgb\_20210218-d8b58813.pth}
  (Last viewed November 29, 2023.)

\bibitem{lopez_2021}
J.A. Lopez, G.~Stemmer, P.~Lopez~Meyer, P.~Singh, J.~Del Hoyo~Ontiveros,
  H.~Cordourier, in \emph{Proceedings of the 6th Detection and Classification
  of Acoustic Scenes and Events 2021 Workshop (DCASE2021)} (Barcelona, Spain,
  2021), pp. 11--15

\bibitem{pytorch}
A.~Paszke, S.~Gross, F.~Massa, A.~Lerer, J.~Bradbury, G.~Chanan, T.~Killeen,
  Z.~Lin, N.~Gimelshein, L.~Antiga, A.~Desmaison, A.~Kopf, E.~Yang, Z.~DeVito,
  M.~Raison, A.~Tejani, S.~Chilamkurthy, B.~Steiner, L.~Fang, J.~Bai,
  S.~Chintala, in \emph{Advances in Neural Information Processing Systems 32},
  ed. by H.~Wallach, H.~Larochelle, A.~Beygelzimer, F.~d\textquotesingle
  Alch\'{e}-Buc, E.~Fox, R.~Garnett (Curran Associates, Inc., 2019), pp.
  8024--8035

\bibitem{agrawal_2020}
V.K. Agrawal, S.S. Maurya, Unsupervised detection of anomalous sounds for
  machine condition monitoring.
\newblock Tech. rep., DCASE2020 Challenge (2020)

\bibitem{python}
G.~Van~Rossum, F.L. Drake, \emph{Python 3 Reference Manual} (CreateSpace,
  Scotts Valley, CA, 2009)

\bibitem{slow_fast}
C.~Feichtenhofer, H.~Fan, J.~Malik, K.~He, Slowfast networks for video
  recognition.
\newblock CoRR \textbf{abs/1812.03982} (2018).
\newblock \urlprefix\url{http://arxiv.org/abs/1812.03982}

\bibitem{mmaction2_2020}
M.~Contributors.
\newblock Openmmlab's next generation video understanding toolbox and
  benchmark.
\newblock \url{https://github.com/open-mmlab/mmaction2} (2020)

\bibitem{adam_optimizer}
D.P. Kingma, J.~Ba.
\newblock Adam: A method for stochastic optimization (2017)

\bibitem{smith_2018}
L.N. Smith, N.~Topin.
\newblock Super-convergence: Very fast training of neural networks using large
  learning rates (2018)

\bibitem{fawcett_2006}
T.~Fawcett, An introduction to roc analysis.
\newblock Pattern Recogn. Lett. \textbf{27}(8), 861–874 (2006)

\end{thebibliography}

\end{document}